\documentclass[journal]{IEEEtran}

\usepackage{fontspec}
\usepackage{amsmath,amssymb,amsfonts}   
\usepackage{graphicx}                   
\usepackage{booktabs}                   
\usepackage{multirow}                   
\usepackage{cite}                       
\usepackage{capt-of}
\usepackage{fancyhdr}                 
\usepackage[ruled,linesnumbered]{algorithm2e} 
\usepackage{xcolor} 
\usepackage[hidelinks]{hyperref} 

\newcommand{\R}{\mathbb{R}}             

\newif\ifarxivprint
\arxivprinttrue

\graphicspath{{figures/}}

\begin{document}

\bstctlcite{BSTcontrol}

\title{PointWalker: Mapless Quadruped Locomotion Directly From Raw LiDAR Point Clouds}

\author{Sicen~Li\textsuperscript{1,2},
        Zhen~Chu\textsuperscript{1,2},
        Chao~Li\textsuperscript{2},
        Jun~Zheng\textsuperscript{4},
        Qiuguo~Zhu\textsuperscript{1},
        and~Jun~Wu\textsuperscript{1,3,*}%
\thanks{\textsuperscript{1}College of Control Science and Engineering,
Zhejiang University, Hangzhou 310027, China (e-mail:
\texttt{\{0625416, 12560199, qgzhu, junwuapc\}@zju.edu.cn}).

\textsuperscript{2}Yunshenchu Technology
Co., Ltd., Hangzhou, China (e-mail: \texttt{lichao@deeprobotics.cn}).

\textsuperscript{3}Zhejiang
Key Laboratory of Additive Manufacturing Technology and Equipment,
Hangzhou, China.

\textsuperscript{4}Hangzhou Public Library, Hangzhou, China (e-mail:
\texttt{zj558@126.com}).

\textsuperscript{*}Corresponding author: Jun Wu.}%
\thanks{Supplementary video is available at {\ttfamily\url{https://youtu.be/QnCFGh7QXIw}}.}%
}

\markboth{IEEE Robotics and Automation Letters. Preprint version.}%
{Li \MakeLowercase{\itshape et al.}: PointWalker: Mapless Quadruped
Locomotion From Raw LiDAR Point Clouds}

\maketitle

\ifarxivprint
\fancypagestyle{arxivfirst}{%
  \renewcommand{\headrulewidth}{0pt}%
  \renewcommand{\footrulewidth}{0pt}%
  \fancyhf{}%
  \fancyfoot[C]{\footnotesize This work has been submitted to the IEEE
  for possible publication. Copyright may be transferred without
  notice, after which this version may no longer be accessible.}}
\thispagestyle{arxivfirst}
\fi

\begin{abstract}
Quadruped robots deployed for inspection and field missions must cross
heterogeneous terrain, from stairs, slopes, and gaps to high
platforms, stepping stones, and balance beams, where safe support is
sparse or see-through. Existing methods interpose a terrain map or a
depth image between sensing and control. Maps are costly to
maintain, corrupted by odometry drift and outliers, and misled by
see-through gratings, while depth cameras degrade with lighting and
surface material. In this paper, we present PointWalker, a mapless
framework whose single reinforcement learning policy directly consumes raw
point clouds from four light detection and ranging (LiDAR) sensors.
Voxelized points are encoded by linear self-attention and fused with
proprioception through cross-attention, so terrain geometry reaches
the policy without maps or foothold estimation. One proximal policy
optimization (PPO) run, with terrain-aware rewards and dual-channel
simulation of see-through gratings, yields one policy for all seven
terrain classes with a 90.1\% mean success rate in simulation.
Deployed zero-shot on a Deep Robotics X30 at 50~Hz onboard, it
crosses 0.8-m gaps, climbs 0.8-m platforms and 45$^{\circ}$
steel-grating stairs, carries a 50-kg payload, and completes a
2.5-km mountain route.
\end{abstract}

\begin{IEEEkeywords}
Legged locomotion, legged robots, LiDAR, point clouds, reinforcement
learning.
\end{IEEEkeywords}

\IEEEpeerreviewmaketitle

\section{Introduction}
\label{sec:intro}

\begin{figure}[!t]
\centering
\includegraphics[width=\columnwidth]{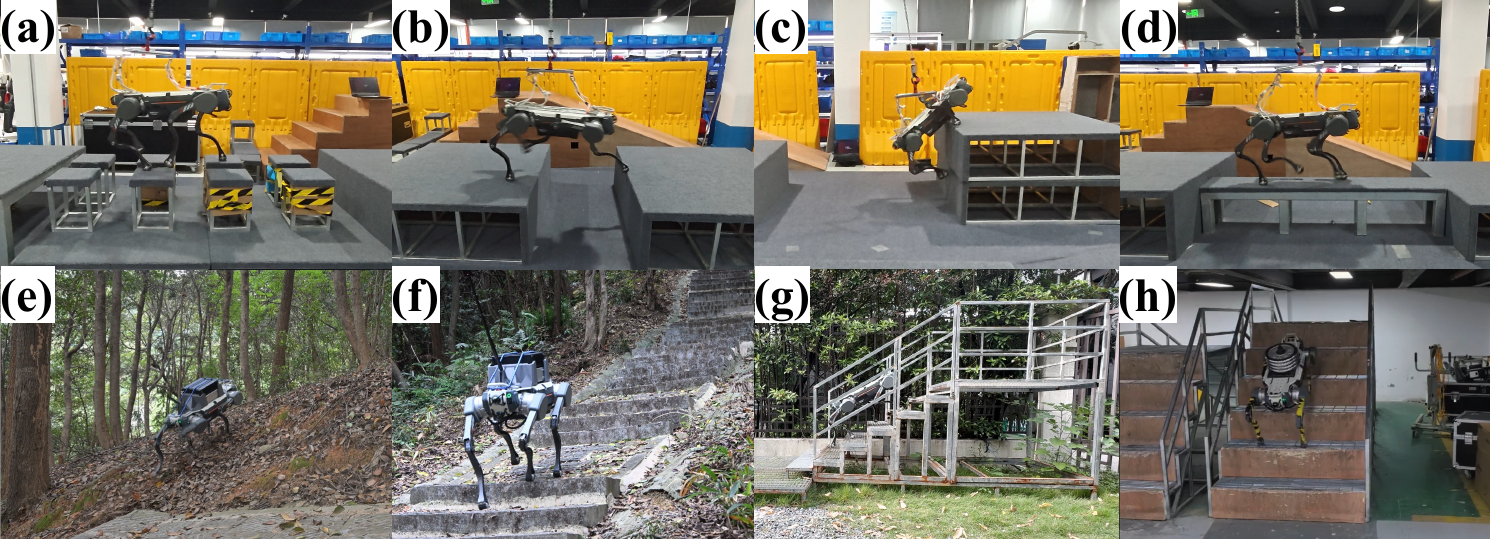}
\caption{Real-world experiments with the X30 quadruped robot, all
driven by one policy. (a)~Stepping stones, (b)~an 80-cm gap, (c)~an
80-cm-high platform, and (d)~a balance beam. (e)~Outdoor mountain
slope and (f)~outdoor stairs on the 2.5-km route (25-kg payload;
Fig.~\ref{fig:outdoor}), (g)~industrial steel-grating stairs, and
(h)~stair ascent with a 50-kg payload.}
\label{fig:teaser}
\end{figure}

Quadruped robots are entering industrial inspection, payload
delivery, and field exploration, where one platform must cross
heterogeneous terrain: stairs, steep slopes, gaps, high platforms,
stepping stones, and balance beams (Fig.~\ref{fig:teaser}). Safe traversal depends on perceiving support
geometry before contact (see-through grating treads, the far edge of
a gap, the next stone in a sparse field), onboard in real time.

Reinforcement learning (RL) dominates quadrupedal locomotion:
massively parallel simulation~\cite{rudin2021parallel} yields robust
proprioceptive policies~\cite{hwangbo2019learning,nahrendra2023dreamwaq},
and teacher--student distillation injects terrain awareness from
privileged height or environment
observations~\cite{lee2020quadrupedal,kumar2021rma}. The resulting policies
are reactive to what the feet touch: on gaps, high platforms, and
sparse footholds (whose decisive geometry lies ahead of and beneath
the body), blind performance collapses.

Perceptive methods therefore insert a terrain representation between
sensing and control. Elevation
maps~\cite{fankhauser2018probabilistic,erni2023mem} serve foothold
planners~\cite{jenelten2020foothold} and perceptive
controllers~\cite{miki2022robust}; depth images and reconstructed
maps are consumed end-to-end by parkour
policies~\cite{zhuang2023parkour,cheng2024extreme,hoeller2024anymal,luo2024pie}.
The burden falls on this interface: odometry drift and registration
error propagate into control, outliers corrupt the map, online
rebuilding consumes onboard compute, see-through gratings violate its
single-height assumption, and depth cameras degrade in bright or dim
light.

Light detection and ranging (LiDAR) provides direct,
illumination-invariant 3-D geometry, yet in locomotion it still
enters the policy through an intermediate form, an elevation
map~\cite{wang2025beamdojo} or a learned
representation~\cite{yuan2026jeplo}, with raw-point input confined
to collision avoidance~\cite{wang2025omni}. Whether a policy can act
on the raw points themselves and span terrain this heterogeneous in
a single run remains open.

In this paper, we propose PointWalker, a mapless perception--control
framework for quadruped locomotion directly from raw LiDAR point
clouds (Fig.~\ref{fig:teaser}). Four LiDARs give a fused
360$^{\circ}$ view of the terrain beneath and ahead; every
control step, the policy consumes 350 voxelized points from the five latest
scans. Linear self-attention encodes the points into tokens, and
cross-attention with proprioceptive history as the query
retrieves task-relevant terrain context: no map, no foothold
estimate. A single proximal policy optimization
(PPO)~\cite{schulman2017ppo} run trains the policy over all terrains,
combining terrain-aware foot-placement rewards, terrain-modulated
takeoff incentives, and a dual-channel simulation that renders
gratings faithfully.

The main contributions of this paper are as follows.
\begin{itemize}
\item \textbf{A mapless LiDAR locomotion interface.} PointWalker is
the first legged controller that learns
terrain traversal directly from raw LiDAR point clouds: no terrain
map, scan projection, world model, or learned latent between sensing
and foot placement; the perception pathway runs in 2.01~ms onboard,
10\% of the 20-ms control period.
\item \textbf{One-stage training of a single policy across seven
terrain classes:} stairs, slopes, discrete obstacles, gaps, high
platforms, stepping stones, and balance beams, with a 90.1\% mean
success rate at the highest curriculum difficulty.
\item \textbf{Zero-shot validation on hardware:} a single
policy on a Deep Robotics X30 traverses 0.8-m gaps, mounts
0.8-m-high platforms, climbs 45$^{\circ}$ steel-grating stairs, keeps
walking with two LiDARs occluded, carries a 50-kg payload (91\% of
body weight), and completes a 2.5-km mountain route with 200~m of
elevation gain.
\end{itemize}

\section{Related Work}
\label{sec:related}

\subsection{Learning-Based Legged Locomotion}
\label{sec:rw_locomotion}

Deep RL is the workhorse of quadrupedal control: Hwangbo
\emph{et al.}~\cite{hwangbo2019learning} learned actuator models
for real ANYmals, massively parallel simulation~\cite{rudin2021parallel}
cut training to minutes, and DreamWaQ~\cite{nahrendra2023dreamwaq}
inferred terrain implicitly from proprioception alone.
Teacher--student pipelines instead grant the teacher privileged
observations and distill a proprioceptive
student~\cite{lee2020quadrupedal,kumar2021rma}; Miki
\emph{et al.}~\cite{miki2022robust} refined the recipe with a
belief state over elevation maps whose trust is gated against
sensor failure, and Ji \emph{et al.}~\cite{ji2022concurrent} trained
the state estimator concurrently with the policy. For terrain whose
support must be seen before it is stepped on, distillation cannot
add information the student never senses; PointWalker trains a
directly perceiving policy in one stage, with privilege confined to
the critic.

\subsection{Perception Interfaces for Agile Locomotion}
\label{sec:rw_perception}

Elevation maps~\cite{fankhauser2018probabilistic,erni2023mem} fuse
depth and point-cloud measurements with odometry into a
robot-centric 2.5-D surface, on which foothold
optimization~\cite{jenelten2020foothold} and attention-based map
encoders~\cite{he2025attention} build. Depth images and maps
reconstructed from multi-sensor point clouds instead feed parkour
policies end-to-end~\cite{zhuang2023parkour,cheng2024extreme,hoeller2024anymal}.
Single-stage frameworks estimate terrain
implicitly~\cite{luo2024pie}, reconstruct local height
maps~\cite{yu2025start}, or predict foothold priors~\cite{wang2026puma}
from vision, with depth-only perception extending to
humanoids~\cite{sun2026dpl}. Specialized lines target hollow
see-through stairs~\cite{tang2026stairmaster}, platform jumps of
varying distance and height~\cite{bellegarda2024robust}, and jumping
over gaps and high platforms~\cite{wang2025sftim}.
Across these systems, the interface rather than the task bounds
performance, and forward depth limits perception to a cone that
excludes the area beneath the body.

\subsection{LiDAR-Based Locomotion}
\label{sec:rw_lidar}

In terrain-aware locomotion, LiDAR reaches the policy in raw form
only for collision avoidance among dynamic
obstacles~\cite{wang2025omni}.
BeamDojo~\cite{wang2025beamdojo} trains humanoid locomotion on
sparse footholds but feeds the policy an elevation map reconstructed
from LiDAR scans; JEPLO~\cite{yuan2026jeplo} learns
mapping-free terrain representations with a joint-embedding
predictive world model, projecting accumulated scans onto a
spherical depth image and acting on the learned latents.
PointWalker removes the last intermediate form: one policy attends
directly to raw voxelized LiDAR points cross-fused with
proprioception (gratings and sparse footholds stay points) and
spans stairs to balance beams in one training run.

\section{Method}
\label{sec:method}

\begin{figure*}[!t]
\centering
\includegraphics[width=\textwidth]{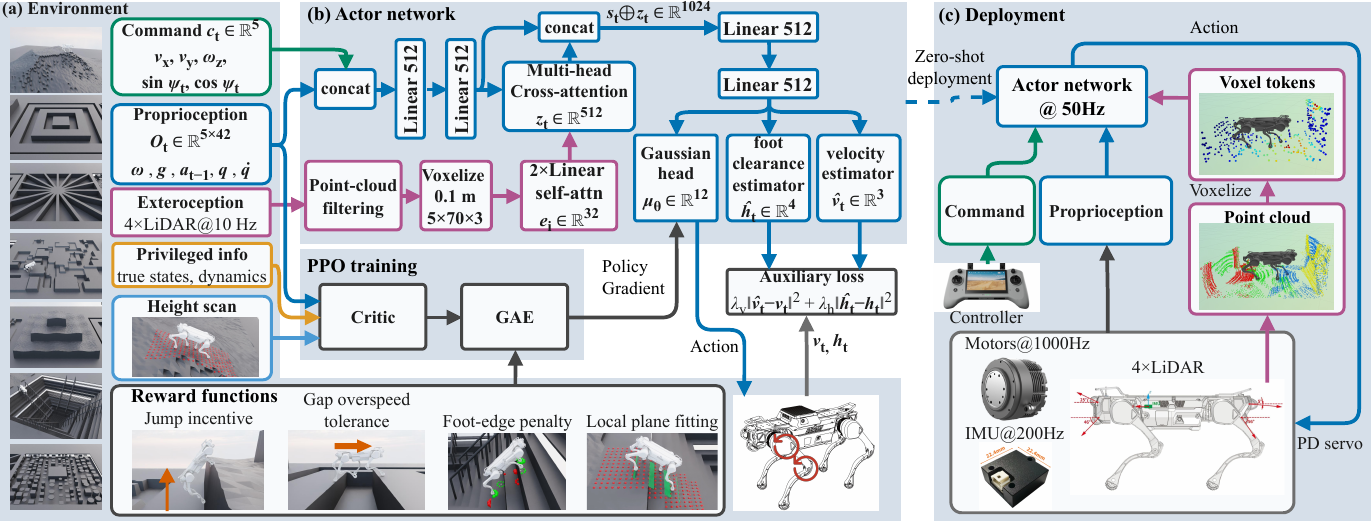}
\caption{PointWalker framework. \textbf{(a)}~Environment: the
policy receives proprioceptive history, commands, and four-LiDAR
point clouds; privileged information (true states, dynamics) feeds
the value network and the terrain-aware rewards in training only.
\textbf{(b)}~Actor network: linear self-attention point encoding,
cross-attention fusion, a Gaussian action head with two auxiliary
heads, and PPO training with an asymmetric critic.
\textbf{(c)}~Deployment: ONNX export and zero-shot 50-Hz inference
on the onboard RK3588.}
\label{fig:framework}
\end{figure*}

Fig.~\ref{fig:framework} illustrates PointWalker: one PPO run in
massively parallel simulation~\cite{mittal2025isaaclab,rudin2021parallel}
produces a single policy covering all seven terrain classes,
transferred zero-shot to the X30.

\subsection{Point Cloud Perception from Onboard LiDAR}
\label{sec:perception}

\textbf{LiDAR configuration.} Four Livox Mid-360 LiDARs are mounted
at the corners of the body, two
at the front and two at the rear, with interleaved pitch, for a
combined 360$^{\circ}$ horizontal field of view and
$-7^{\circ}$--$52^{\circ}$ vertical range. Simulation reproduces this
geometry by casting multiple rays that also hit the perception-only
structural layer (Section~\ref{sec:training}) and the thigh and shank
meshes to model leg self-occlusion. The rays are randomly
permuted into five groups emitted in rotation (an effective 10-Hz
scan rate), mimicking the Mid-360's non-repetitive scanning. At 50~Hz, one group is read per step and the latest five
retained: $5\times70=350$ points spanning 0.1~s.

\textbf{Preprocessing and degradation.} Each step, the hits are
transformed to the base frame, and points within
0.15~m of a sensor origin or inside the body bounding box are
removed. Gaussian noise ($\sigma=0.02$~m) is added; the
rest is voxelized at 0.1-m resolution over the perception region
$x\in[-1.2,1.2]$, $y\in[-0.6,0.6]$, $z\in[-1.0,0.5]$~m (the tops of
the tallest curriculum platforms rise above this bound; their front
faces remain in view) with per-voxel averaging; and 70 points per
group are sampled from
occupied voxels into the fixed-size tensor
$\mathcal{P}_t\in\R^{5\times70\times3}$. Training also
injects sensor degradation (2\% invalidated points, 2\% random
outliers, 40--60-ms latency), so the policy trains against an
already-degraded sensing chain. Invalid points are masked, not
removed, in the attention below. Fixed-size
sampling keeps onboard compute independent of scene complexity.

\textbf{Linear-attention encoding.} Each
slot, base-frame coordinates plus a temporal scan-group encoding
(4-D input), is linearly embedded to 32 dimensions and
contextualized by two levels of linear
self-attention~\cite{katharopoulos2020transformers}, yielding 32-D
point tokens
\begin{equation}
\boldsymbol{e}_i=\frac{\sum_{j}\phi\!\left(\boldsymbol{W}_Q\boldsymbol{x}_i\right)^{\!\top}\phi\!\left(\boldsymbol{W}_K\boldsymbol{x}_j\right)\boldsymbol{x}_j}{\sum_{j}\phi\!\left(\boldsymbol{W}_Q\boldsymbol{x}_i\right)^{\!\top}\phi\!\left(\boldsymbol{W}_K\boldsymbol{x}_j\right)},\quad
\phi(\boldsymbol{x})=\mathrm{elu}(\boldsymbol{x})+\mathbf{1},
\label{eq:linattn}
\end{equation}
where $\boldsymbol{x}_i$ is the 32-D embedding of the $i$th point,
$\boldsymbol{W}_Q$ and $\boldsymbol{W}_K$ the query and key
projections, and $\phi$ a positive feature map, with
$\boldsymbol{x}_j$ as its own value vector and the denominator
clamped for numerical stability. Linear attention
scales with the number of points instead of quadratically, avoiding
the onboard latency of standard attention. Unlike shared-MLP
encoders~\cite{qi2017pointnet}, it captures inter-point
geometry (the parallel bars of a grating) that identifies
steppable regions.

\subsection{Policy Network With Cross-Attention Fusion}
\label{sec:policy}

\textbf{Observations and commands.} The single-frame
observation is
$\boldsymbol{o}_t=\left[\tfrac{1}{4}\,\boldsymbol{\omega}_t^{\top},\ %
\boldsymbol{g}_t^{\top},\ \boldsymbol{a}_{t-1}^{\top},\ %
\boldsymbol{q}_t^{\top},\ \tfrac{1}{10}\,\dot{\boldsymbol{q}}_t^{\top}\right]^{\top}\in\R^{42}$,
where $\boldsymbol{\omega}_t$ is the base angular velocity,
$\boldsymbol{g}_t$ the projected gravity,
$\boldsymbol{a}_{t-1}$ the previous action, and $\boldsymbol{q}_t$,
$\dot{\boldsymbol{q}}_t$ the positions and velocities of the 12
joints. Base linear velocity, not measurable without external
localization, is excluded; channels carry Gaussian noise, mounting
misalignment, and delays. The latest five
frames (0.1~s) stack into
$\boldsymbol{O}_t\in\R^{5\times42}$. The command vector
$\boldsymbol{c}_t=[v_x^{\mathrm{cmd}},\,v_y^{\mathrm{cmd}},\,\omega_z^{\mathrm{cmd}},\,\sin\psi_t,\,\cos\psi_t]\in\R^{5}$
combines velocity commands ($\pm1.5$~m/s forward, $\pm0.9$~m/s
lateral, $\pm1$~rad/s yaw rate) with the phase of a 1.4-Hz gait
clock as the rhythmic prior for
swing--stance alternation~\cite{margolis2022walk}.

\textbf{Cross-attention fusion.} A temporal branch concatenates each
frame of $\boldsymbol{O}_t$ with
$\boldsymbol{c}_t$, appends frame position encodings, and encodes
the flattened vector with a two-layer 512-D multilayer perceptron
(MLP; linear, layer normalization, sigmoid linear unit) into
$\boldsymbol{s}_t$. The fusion module
then queries the point tokens with this feature through multi-head
cross-attention (MHA; Fig.~\ref{fig:framework}(b)):
\begin{equation}
\boldsymbol{z}_t=\mathrm{MHA}\!\left(\boldsymbol{u}_t,\ %
\{\boldsymbol{e}_i\}_{i=1}^{350}\right)\in\R^{512},
\label{eq:fusion}
\end{equation}
where the query $\boldsymbol{u}_t$ is a per-head projection of
$\boldsymbol{s}_t$ and 16 heads of 32 dimensions attend over the 350
tokens with invalid points masked and concatenate into the 512-D
terrain context $\boldsymbol{z}_t$. Querying with proprioception and
commands retrieves task-relevant geometry (the gap edge in the
crossing direction, the support near a swinging leg) and suppresses
the rest.

\textbf{Action and auxiliary heads.} The backbone concatenates
$\boldsymbol{s}_t$ with $\boldsymbol{z}_t$
(1024-D), passes through a two-layer 512-D MLP, and feeds three
linear heads. The action head parameterizes the diagonal Gaussian
$\pi_\theta(\cdot\mid\mathcal{P}_t,\boldsymbol{O}_t,\boldsymbol{c}_t)=\mathcal{N}(\boldsymbol{\mu}_\theta,\mathrm{diag}\,\boldsymbol{\sigma}_\theta^{2})$
with mean $\boldsymbol{\mu}_\theta\in\R^{12}$ and log standard
deviation clipped to $[-6,2]$. Two auxiliary heads estimate the
base linear velocity $\hat{\boldsymbol{v}}_t\in\R^{3}$ and foot
clearances $\hat{\boldsymbol{h}}_t\in\R^{4}$ under mean-squared-error
supervision \eqref{eq:loss}. The supervision forces the backbone to
encode the
motion state and local terrain geometry that blind policies recover
implicitly and dynamical drift
corrupts~\cite{ji2022concurrent,miki2022robust}; unlike a separate
concurrently trained estimator~\cite{ji2022concurrent}, the heads
add no extra forward pass and yield online estimates.

\textbf{Action mapping and joint control.} The action (sampled
during training, the mean
$\boldsymbol{\mu}_\theta$ at deployment) is scaled and clipped to
joint position targets
$\boldsymbol{q}_t^{\mathrm{des}}=\operatorname{clip}\!\left(\boldsymbol{q}^{0}+0.25\,\boldsymbol{a}_t,\ %
\boldsymbol{q}_{\lim}^{-},\ \boldsymbol{q}_{\lim}^{+}\right)$ around
the default standing pose
$\boldsymbol{q}^{0}=[0,-1,1.8]$~rad per leg. Joint
proportional--derivative (PD) controllers track these targets with
hip-abduction/hip-pitch/knee gains
$K_p=100/100/120$~N\,m/rad, $K_d=2.5/2.5/3$~N\,m\,s/rad, and torque
limits $84/84/150$~N\,m, without additional low-level filtering.

\textbf{Asymmetric actor--critic.} Beyond commands and actions, the
asymmetric critic~\cite{pinto2018asymmetric} reads training-only privileged
inputs through a multilayer 512-D MLP: four frames of
noise-free true state (base velocity, joint states, foot contacts),
a height map around the body, and the ground-truth
randomized dynamics parameters
(Table~\ref{tab:hyper}). Privilege
is confined to the value pathway and absent from the exported model.
Deployment thus needs no height map or physical estimates and avoids
the distillation gap of two-stage teacher--student
pipelines~\cite{lee2020quadrupedal,kumar2021rma}. The
policy minimizes the PPO loss jointly with the value and
auxiliary losses:
\begin{equation}
\begin{aligned}
L(\theta,\vartheta)=\ &-\,\hat{\mathbb{E}}\!\left[\min\!\left(\eta_t\hat{A}_t,\ %
\operatorname{clip}\!\left(\eta_t,1{-}\epsilon,1{+}\epsilon\right)\hat{A}_t\right)\right]\\
&+\,\beta_v\,\hat{\mathbb{E}}\!\left[\left(V_{\vartheta}(\boldsymbol{\xi}_t){-}V^{\mathrm{targ}}\right)^{2}\right]
+\lambda_v\|\hat{\boldsymbol{v}}_t{-}\boldsymbol{v}_t\|^{2}\\
&+\lambda_h\|\hat{\boldsymbol{h}}_t{-}\boldsymbol{h}_t\|^{2},
\end{aligned}
\label{eq:loss}
\end{equation}
where $\eta_t$ is the action probability ratio, $\hat{A}_t$ the
generalized advantage estimate, $V_{\vartheta}$ the value network
with parameters $\vartheta$ and input $\boldsymbol{\xi}_t$,
$\epsilon=0.2$ the clip parameter, $\beta_v=1.0$ and
$\lambda_v=\lambda_h=0.1$ the value and auxiliary coefficients, and
$\boldsymbol{v}_t$, $\boldsymbol{h}_t$ the privileged ground truth;
remaining hyperparameters follow Section~\ref{sec:training}.

\subsection{Terrain-Aware Reward Design}
\label{sec:reward}

\textbf{Reward composition.} The per-step reward is the weighted sum
$r_t=\sum_{i}w_i r_i$;
Table~\ref{tab:reward} lists all terms, weights, and equations,
with tracking terms from~\cite{nahrendra2023dreamwaq}. In
Table~\ref{tab:reward},
$\tau_k$, $\dot{q}_k$, $\ddot{q}_k$ are joint torque, velocity, and
acceleration; $F_j$ the contact force of foot $j$; $b_j$ the contact
flag; $\boldsymbol{p}_{f,j}$ the foot position and
$\boldsymbol{p}_{h,j}$ its nominal standing location; sw/st mark
swing and stance. Rewards use noise-free
simulator state; the policy never observes base velocity, whose
estimation rests with the auxiliary heads and value network
(Section~\ref{sec:policy}).

\begin{table}[!t]
\caption{Reward Terms and Weights}
\label{tab:reward}
\centering
\scriptsize
\renewcommand{\arraystretch}{0.95}
\setlength{\tabcolsep}{1.5pt}
\begin{tabular}{@{}llc@{}}
\toprule
Reward term & Equation & Weight\\
\midrule
Lin.\ vel.\ tracking & $\exp(-4\|\boldsymbol{v}^{xy,\mathrm{cmd}}-\boldsymbol{v}^{xy}\|^{2})$ & $+1.1$\\
Ang.\ vel.\ tracking & $\exp(-4(\omega_z^{\mathrm{cmd}}-\omega_z)^{2})$ & $+0.6$\\
Lin.\ velocity (z) & $v_z^{2}$ & $-2.0$\\
Ang.\ velocity (xy) & $\|\boldsymbol{\omega}^{xy}\|^{2}$ & $-0.06$\\
Body height & $(h^{\mathrm{des}}-h^{\mathrm{b}})^{2}$ & $-5.0$\\
Terrain alignment & $1-\exp(-\|\boldsymbol{g}-\boldsymbol{n}_t\|^{2})$ & $-1.0$\\
Collision & $N_{\mathrm{c}}$ & $-0.2$\\
Joint limits & $\sum\nolimits_k (q_k-\operatorname{clip}(q_k,q_k^{-},q_k^{+}))^{2}$ & $-0.2$\\
Stumble & $\sum\nolimits_{j}\mathbf{1}[\|\boldsymbol{F}_j^{xy}\|{>}5|F_j^{z}|]$ & $-0.1$\\
Joint power & $\sum\nolimits_k |\tau_k\dot{q}_k|$ & $-0.0015$\\
Torques & $\sum\nolimits_k \tau_k^{2}$ & $-0.04$\\
Joint accelerations & $\sum\nolimits_k \ddot{q}_k^{2}$ & $-0.001$\\
Landing impact & $\sum\nolimits_j b_j\|\boldsymbol{v}_{f,j}\|^{2}$ & $-0.15$\\
Foot slip & $\sum\nolimits_{j\in\mathrm{st}}\|\boldsymbol{v}_{f,j}^{xy}\|^{2}$ & $-0.15$\\
Foot placement & $\sum\nolimits_j\|\boldsymbol{p}_{f,j}-\boldsymbol{p}_{h,j}\|^{2}$ & $-1.0$\\
Footswing height & $\sum\nolimits_{j\in\mathrm{sw}}(h_{f,j}-h^{\mathrm{bez}})^{2}$ & $-10.0$\\
Phase tracking & $\sum\nolimits_j \mathbf{1}[j{\in}\mathrm{sw}](1{-}b_j)+\mathbf{1}[j{\in}\mathrm{st}]\exp(-\|\boldsymbol{v}_{f,j}^{xy}\|^{2})$ & $+0.15$\\
Foot-edge penalty & $\frac{1}{M}\sum\nolimits_{j\in\mathrm{st}}\sum\nolimits_{m}\mathbf{1}[d_{j,m}\notin\mathcal{W}]$ & $-0.5$\\
Jump incentive & $v_z\,(h^{\mathrm{p}}-h_0)^{+}$, $h_0=0.25$\,m & $+2.0$\\
Action rate & $\|\boldsymbol{a}_t-\boldsymbol{a}_{t-1}\|^{2}$ & $-0.04$\\
Smoothness & $\|\boldsymbol{a}_t-2\boldsymbol{a}_{t-1}+\boldsymbol{a}_{t-2}\|^{2}$ & $-0.04$\\
\bottomrule
\end{tabular}
\end{table}

\textbf{Local plane fitting.} One reward basis covers all terrains:
a least-squares plane
fitted to the privileged height samples around the body yields the
terrain normal $\boldsymbol{n}_t=[A,B,-1]/\|[A,B,-1]\|$ of the fitted
plane $z{=}Ax{+}By{+}D$ and the body height $h^{\mathrm{b}}$ above it
($h^{\mathrm{des}}$ the nominal
standing height). The alignment term penalizes the deviation of the
projected gravity from $\boldsymbol{n}_t$ rather than the world
vertical, so posture conforms to slopes and steps with no
per-terrain switching.

\textbf{Foot edge and swing trajectory.} Landing quality is scored
by casting $M=11$ privileged downward rays within a disc of
radius 0.085~m under each foot (the privileged ring scan). For a stance
foot $j$, $d_{j,m}$ (the vertical distance from the $m$th hit to
the foot reference plane) is valid support only inside
$\mathcal{W}=[-0.11,0.02]$~m (misses, edge overhangs, and
penetrating protrusions are invalid). The foot-edge penalty
charges the invalid fraction (Table~\ref{tab:reward}); related
penalties appear
in~\cite{cheng2024extreme,yu2025start,zhang2026rpl}. Swing
feet track a quadratic B\'ezier profile $h^{\mathrm{bez}}$.

\textbf{Dynamic traversal incentives.} We add two terrain-modulated
incentives for the pre-takeoff momentum
that gaps and high platforms demand but purely tracking-style
rewards suppress~\cite{zhuang2023parkour,wang2025sftim}.
For gaps, the commanded
speed along the crossing direction is boosted by
$\kappa(G_w-G_0)^{+}$ with the overspeed tolerance relaxed ($G_w$ the
gap width ahead, $\kappa=2$~s$^{-1}$, $G_0=0.3$~m). For platforms,
the jump incentive rewards the product $v_z(h^{\mathrm{p}}-h_0)^{+}$ of vertical velocity and
excess height ($h^{\mathrm{p}}$ the platform
height ahead, $h_0=0.25$~m). Both are estimated online
from the privileged height samples (gap width as the width of
support-free columns, platform height as the difference between the
means of the highest and lowest point sets). The incentive scales
with difficulty.

\textbf{Termination.} Episodes terminate on persisting undesirable
body contact
($\approx$1~s), tipping over, sinking below the support surface,
falling into a gap, leaving the terrain, or timeout (20~s); failures
and timeouts enter the advantage estimation differently.

\subsection{Dual-Channel Simulation and One-Stage Training}
\label{sec:training}

\begin{table}[!t]
\caption{Difficulty Curriculum of the Seven Training Terrains}
\label{tab:terrain}
\centering
\scriptsize
\setlength{\tabcolsep}{2pt}
\begin{tabular}{@{}lll@{}}
\toprule
Terrain & Difficulty parameter & Curriculum range\\
\midrule
Stairs & Step height (tread $0.28$\,m) & $0.1\to0.28$\,m\\
Slopes & Slope ($\tan\theta$) & $0.2\to0.65$\\
Discrete obstacles & Obstacle height & $0.1\to0.23$\,m\\
Gaps & Gap width & $0.15\to1.1$\,m\\
High platforms & Platform height & $0.25\to1.2$\,m\\
Stepping stones & Stone spacing & $0.1\to0.55$\,m\\
 & Stone width & $0.8\to0.35$\,m\\
Balance beams & Beam width & $0.9\to0.3$\,m\\
\bottomrule
\end{tabular}
\end{table}

\begin{table}[!t]
\caption{Training and Domain Randomization Configuration}
\label{tab:hyper}
\centering
\scriptsize
\renewcommand{\arraystretch}{0.95}
\setlength{\tabcolsep}{2pt}
\begin{tabular}{@{}ll@{}}
\toprule
Parameter & Value\\
\midrule
\multicolumn{2}{@{}l}{\textbf{Training configuration}}\\
Parallel environments & 8192\\
GPUs / training time & $4\times$ RTX 5090 / $\sim$50\,h\\
Optimizer / learning rate & Adam / $3\times10^{-4}$\\
Discount $\gamma$ / GAE $\lambda$ & 0.99 / 0.95\\
Clip $\epsilon$ / grad.\ clip / epochs & 0.2 / 1.0 / 5\\
\midrule
\multicolumn{2}{@{}l}{\textbf{Domain randomization}}\\
Friction / restitution & $\mathcal{U}(0.3,1.2)$ / $\mathcal{U}(0,0.3)$\\
Link mass / PD gains & $\mathcal{U}(0.85,1.15)$ / $\mathcal{U}(0.9,1.1)$\\
CoM displacement & $x\pm0.1$, $y\pm0.05$, $z\in(-0.075,0.15)$\,m\\
Payload & $\mathcal{U}(-10,50)$\,kg\\
Motor strength & $\mathcal{U}(0.85,1.15)$\\
Action delay / IMU orientation & $0$--$28$\,ms / $\pm0.1$\,rad\\
Observation noise & Joint pos.\ $\pm0.01$\,rad, joint vel.\ $\pm1.5$\,rad/s,\\
 & base ang.\ vel.\ $\pm0.2$\,rad/s, proj.\ gravity $\pm0.1$\\
Point cloud degradation & $\sigma{=}0.02$\,m noise, $2\%$ invalid/outlier,\\
 & $40$--$60$\,ms latency\\
\bottomrule
\end{tabular}
\end{table}

\textbf{Physics and perception channels.} The physics channel holds
the collidable height field and meshes;
privileged height sampling and the foot rays of
Section~\ref{sec:reward} cast only against it. The perception channel
is a structural layer that enters LiDAR casting but never in
collision:
steel-grating treads and risers (5-mm flat bars at 40-mm pitch),
stair handrails, random posts, hanging bars, and platform edge beams and posts. The privileged side sees solid support,
while the policy's point cloud sees the sparse grating,
handrails, and clutter of a real industrial stair, flat bars and
spacing preserved as points. The policy learns to land on solid
support while perceiving the see-through structure.

\textbf{Terrain curriculum.} The seven terrains
(Table~\ref{tab:terrain}) form a
curriculum of $10\times10$-m tiles, robots starting on random
difficulty rows; discrete
obstacles are used in simulation only. Following
legged\_gym~\cite{rudin2021parallel}, traveling beyond half the tile
length promotes a robot to a harder row, falling short of half the
commanded distance demotes it. Balance beams are narrow beams
arranged in a star.

\textbf{Domain randomization.} To bridge the reality gap, training
randomizes contact, mass,
actuation, and sensing~\cite{peng2017dynamics}; sensing-side
degradation follows Section~\ref{sec:perception}.
Half of the environments apply left--right mirror augmentation
(joint reordering with matching reflections of observations,
commands, actions, and points). Training runs in Isaac Lab with
massive parallelism; the full configuration is in
Table~\ref{tab:hyper}.

\textbf{Deployment.} Only the policy network is exported to ONNX
(Fig.~\ref{fig:framework}(c)) and runs deterministically (action
mean) at 50~Hz on the X30's onboard RK3588. The
cross-attention weights are read out with the action and projected
onto the points as per-point importance for interpreting foothold
decisions.

\section{Experiments}
\label{sec:exp}

\setlength{\abovecaptionskip}{1pt plus 0.5pt minus 0.5pt}
\setlength{\belowcaptionskip}{1pt plus 0.5pt minus 0.5pt}
\setlength{\textfloatsep}{2.5pt plus 1pt minus 1pt}
\setlength{\floatsep}{2.5pt plus 1pt minus 1pt}
\setlength{\intextsep}{2.5pt plus 1pt minus 1pt}

We report the experimental setup (Section~\ref{sec:setup}),
simulation results (Section~\ref{sec:sim}), and hardware results
(Section~\ref{sec:real}); discrete obstacles are evaluated in
simulation only.

\subsection{Experimental Setup}
\label{sec:setup}

\begin{table}[!t]
\caption{Compute Cost of the Perception Representations}
\label{tab:compute}
\centering
\scriptsize
\renewcommand{\arraystretch}{0.95}
\setlength{\tabcolsep}{1.5pt}
\begin{tabular}{@{}lccc@{}}
\toprule
Representation & Params (M) & FLOPs (M) & Latency (ms)\\
\midrule
\begin{tabular}[t]{@{}l@{}}PointWalker, linear attn.,\\ 0.1-m voxels\end{tabular} & 1.47 & 7.35 & 2.01$\,\pm\,$0.85\\
w/ Standard attention & 1.47 & 22.67 & 8.23$\,\pm\,$3.68\\
w/ 0.05-m voxels & 1.47 & 19.52 & 7.33$\,\pm\,$2.40\\
w/ 0.2-m voxels & 1.47 & 4.36 & 1.03$\,\pm\,$0.45\\
Depth-image encoding & 2.16 & 11.82 & 4.50$\,\pm\,$2.25\\
Height-map encoding & 1.26 & 2.90 & 1.18$\,\pm\,$0.43\\
Blind & 0.75 & 1.89 & 0.38$\,\pm\,$0.06\\
\bottomrule
\end{tabular}
\end{table}

\textbf{Platform and deployment.} The robot is the Deep Robotics X30
quadruped (55~kg, 0.9-m body, 12 actuated joints); sensing and
deployment follow
Sections~\ref{sec:perception} and~\ref{sec:training}.
Table~\ref{tab:compute} reports the onboard cost of each perception
representation: network parameters, FLOPs, and full-pathway latency
including voxelization, across 100 ONNX Runtime passes at batch size 1
on the RK3588 CPU. Linear attention with 0.1-m voxels holds the per-step
latency at 2.01~ms, 10\% of the 20-ms control period, leaving headroom
for servo control and safety monitoring.

\begin{figure}[!t]
\centering
\includegraphics[width=\columnwidth]{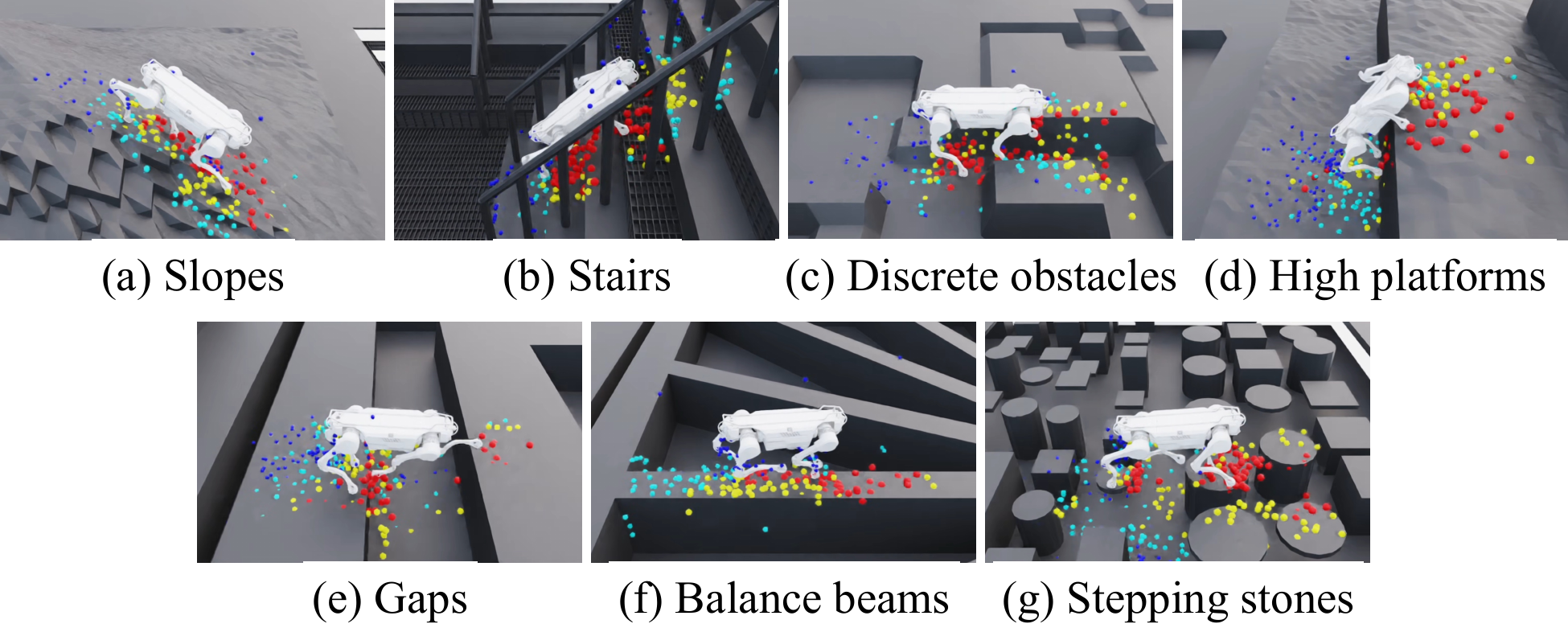}
\caption{Snapshots of the seven training terrains in simulation
(Table~\ref{tab:terrain}); points colored by cross-attention weight
(Section~\ref{sec:policy}): red = high, blue = low.}
\label{fig:simterrains}
\end{figure}

\textbf{Metrics.} A simulation trial succeeds if the robot crosses
the full 10-m tile within the 20-s timeout without a failure
termination (Section~\ref{sec:reward}). We evaluate success rates
(Table~\ref{tab:sim}) at the highest curriculum difficulty
(Table~\ref{tab:terrain}; Fig.~\ref{fig:simterrains}) with a 1.0-m/s
command, the structural layer enabled, training-level sensor
degradation, and the deterministic mean policy, matching deployment.
Each cell covers 200 trials over 5 seeds. Landing quality is the edge-landing ratio,
the invalid fraction of the privileged ring scan
(Section~\ref{sec:reward}), on stairs, balance beams, and stepping
stones.

\textbf{Baselines and ablations.} Interface-level contrasts swap
only the terrain representation (Height-map encoding) or remove
perception altogether (Blind); the external-method re-run
(Depth-image encoding) keeps the source methods' perception and
adopts our terrain, curriculum, and environments. Published hardware
numbers are not comparable across platforms
(Section~\ref{sec:related}), and re-running external systems on
hardware is precluded by equipment and safety; method-level
comparison therefore stays in simulation. All entries are retrained
as single policies under
Table~\ref{tab:hyper} with only the ablated component changed; the
simulation-evaluated entries share five seeds and one protocol, the
rest being verified qualitatively on hardware
(Section~\ref{sec:real}): \textbf{Depth-image encoding} follows the
egocentric depth inputs of PIE~\cite{luo2024pie} and
START~\cite{yu2025start}, virtual-camera rendered as in the
originals (Tables~\ref{tab:compute},~\ref{tab:sim});
\textbf{Height-map encoding} aggregates the 350-point tensor into a
0.1-m 2.5-D height map encoded by an
MLP~\cite{fankhauser2018probabilistic,lee2020quadrupedal}, the
deployed form of elevation-map methods~\cite{wang2025beamdojo}
(Tables~\ref{tab:compute},~\ref{tab:sim}); \textbf{Blind} removes the
point branch~\cite{nahrendra2023dreamwaq}
(Tables~\ref{tab:compute},~\ref{tab:sim}); \textbf{w/o Dual-channel
terrain} removes the perception-only structural layer
(Section~\ref{sec:training}; hardware, qualitative); \textbf{w/o
Foot-edge penalty} removes the foot-edge penalty
(Table~\ref{tab:reward}; Fig.~\ref{fig:edge}; hardware,
qualitative); \textbf{w/o Overspeed tolerance} removes the gap speed
boost and overspeed tolerance (Section~\ref{sec:reward}; hardware,
qualitative); \textbf{w/o Jump incentive} removes the jump incentive
(Table~\ref{tab:reward}; hardware, qualitative); \textbf{w/o
Cross-attn.\ fusion} replaces cross-attention \eqref{eq:fusion} with
mean pooling and concatenation (Table~\ref{tab:sim}); and \textbf{w/
Standard attention / 0.05-m / 0.2-m voxels} uses softmax attention
instead of linear, or changes the voxel size
(Tables~\ref{tab:compute},~\ref{tab:sim}).

\subsection{Simulation Results}
\label{sec:sim}

\begin{table}[!t]
\caption{Simulation Success Rates (\%)}
\label{tab:sim}
\centering
\scriptsize
\renewcommand{\arraystretch}{0.95}
\setlength{\tabcolsep}{1.2pt}
\begin{tabular}{@{}lcccccccc@{}}
\toprule
Method & Stairs & Slopes & Obst. & Gaps & Plat. & Stones & Beams & Mean\\
\midrule
\textbf{PointWalker (full)} & 93.2 & 96.3 & 93.2 & 86.1 & 85.3 & 83.8 & 92.7 & 90.1\\
\midrule
Depth-image enc. & 94.6 & 94.9 & 91.7 & 71.5 & 82.9 & 68.4 & 89.3 & 84.8\\
Height-map enc. & 86.7 & 97.2 & 93.3 & 82.4 & 87.1 & 75.8 & 89.1 & 87.4\\
Blind & 82.6 & 84.1 & 90.3 & 18.5 & 15.2 & 6.4 & 11.7 & 44.1\\
\midrule
w/o Cross-attn. & 82.4 & 92.1 & 89.6 & 74.8 & 75.2 & 68.9 & 77.5 & 80.1\\
w/ Standard attn. & 94.3 & 95.7 & 94.0 & 87.2 & 84.8 & 84.7 & 92.1 & 90.4\\
w/ 0.05-m voxels & 92.8 & 97.1 & 94.5 & 88.5 & 87.4 & 90.1 & 94.3 & 92.1\\
w/ 0.2-m voxels & 82.7 & 95.6 & 92.5 & 79.1 & 79.6 & 67.7 & 78.3 & 82.2\\
\bottomrule
\end{tabular}
\end{table}

\textbf{Overall traversal and baselines.} PointWalker averages 90.1\%
across the seven terrains and never drops below 83.8\%, with one
policy and no per-terrain switching (Table~\ref{tab:sim}).
Height-map encoding trails on stairs and stepping stones (86.7\% and
75.8\% versus 93.2\% and 83.8\%): on the see-through grating
(Fig.~\ref{fig:simterrains}(b)), the 2.5-D aggregation yields
spurious heights that are neither safe support nor true geometry, and
rasterization erases the geometry of sparse footholds. Slopes and
platforms score slightly higher, yet the mean trails by 2.7 points.
Blind keeps 82.6\% and
84.1\% on stairs and slopes, terrain the feet can probe, but
collapses below 30\% on gaps, platforms, stones, and
beams, where support must be seen before contact. Depth-image
encoding averages 84.8\%: gaps and stones fall well short (71.5\% and
68.4\% versus 86.1\% and 83.8\%) and platforms and beams slightly
short, as the forward camera cone, an interface property rather than
an encoding deficit, misses the area beneath and beside the body and
forces reliance on network memory. On stairs, slopes, and
discrete
obstacles the two trade places (94.6\%, 94.9\%, 91.7\%). Image space
keeps the grating's 2-D bars that the height map loses on stairs,
at $2.2\times$ the latency (Table~\ref{tab:compute}).

\begin{figure}[!t]
\centering
\includegraphics[width=\columnwidth]{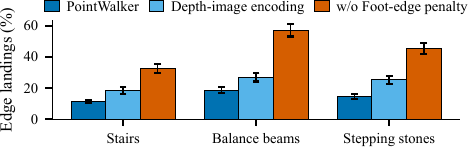}
\caption{Edge-landing ratio (invalid fraction of the privileged ring
scan, Section~\ref{sec:setup}) on the three foothold-restricted
terrains; error bars: standard deviation across five seeds.}
\label{fig:edge}
\end{figure}

\textbf{Component ablation.} Fig.~\ref{fig:edge} reports the
edge-landing ratio: PointWalker stays lowest (11.5\%--18.7\%);
Depth-image encoding rises to 18.6\%--26.9\%, about 60\% higher on
average and largest on stepping stones; w/o Foot-edge penalty reaches
32.4\%--57.2\%, roughly three
times PointWalker's. Privileged landing supervision thus converts
directly into an edge-landing constraint; the other three components
follow in Section~\ref{sec:real}.

\begin{figure}[!t]
\centering
\includegraphics[width=\columnwidth]{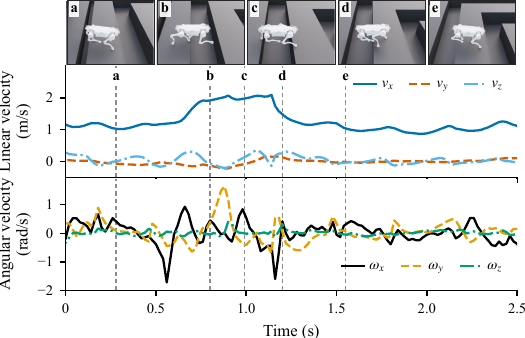}
\caption{Gap crossing with the overspeed tolerance (simulation). Top:
keyframes a--e in time order; middle and bottom: base-frame linear
and angular velocities, dashed lines marking the keyframe instants.}
\label{fig:gap}
\end{figure}

\textbf{Gap-crossing dynamics.} In Fig.~\ref{fig:gap}, the robot
cruises at $v_x\approx1.1$~m/s, within the soft tracking tolerance
of the 1.0-m/s command (a), accelerates into the near edge as the
front legs reach across it (b), peaks at 2.1~m/s, 90\% above
cruise, over the gap (c, d). $\omega_y$ swings to 1.6~rad/s, and the robot
resumes cruising on the far side (e). The pre-edge acceleration is
what the overspeed tolerance permits (Section~\ref{sec:reward}).

\begin{figure}[!t]
\centering
\includegraphics[width=\columnwidth]{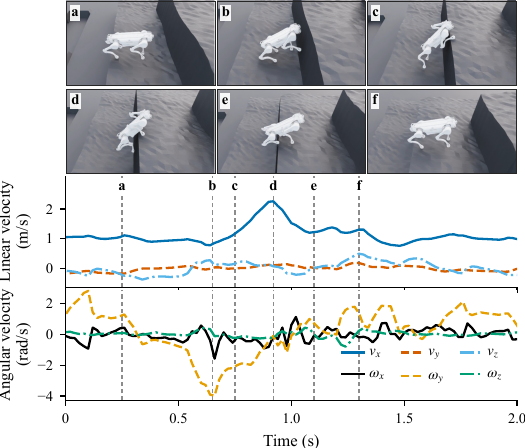}
\caption{Climbing onto a high platform with the jump incentive
(simulation). Top: keyframes a--f in time order; middle and bottom:
base-frame velocities with the keyframe instants marked.}
\label{fig:platform}
\end{figure}

\textbf{Platform-mounting dynamics.} In Fig.~\ref{fig:platform}, the
robot cruises at 1.0~m/s (a), accelerates at the wall, rears up with
$\omega_y$ up to 4~rad/s (b--c), and peaks at 2.3~m/s, more than
twice the cruise, while hooking the top edge and pivoting over it
(c--e). A 0.5-m/s upward $v_z$ pulse marks the hind-leg push onto
the top (e, f). The build-up and the vertical burst are what the
jump incentive $v_z(h^{\mathrm{p}}-h_0)^{+}$ (Table~\ref{tab:reward})
elicits.

\textbf{Architecture variants.} Mean pooling instead of
cross-attention costs 10 points on average (80.1\% versus
90.1\%): slopes and discrete obstacles degrade least (92.1\%, 89.6\%),
stepping stones and beams most (68.9\%, 77.5\%). Decisive foothold
points are a small fraction of the input, retrieved only by a
motion-conditioned query (Section~\ref{sec:policy}). Standard softmax
attention averages 90.4\%, within 1.1 points per column, at
$4.1\times$ the latency (Table~\ref{tab:compute}): linear attention
costs less than half a point and keeps onboard headroom. Voxels of
0.05~m add 2.0 points (92.1\%, stones $+6.3$) at $3.6\times$ latency.
Voxels of 0.2~m lose 7.9 (82.2\%), sparing slopes and discrete obstacles
but hitting stones ($-16.1$), beams ($-14.4$), and stairs ($-10.5$),
whose narrow supports and grating bars approach the resolution limit.
The voxel variants' FLOPs (Table~\ref{tab:compute}) track the point
count, not the parameters: points are sampled from occupied voxels, so
finer voxels admit more points and coarser voxels fewer.
The 0.1-m voxel balances accuracy and compute.

\subsection{Real-World Results}
\label{sec:real}

We deploy the single policy of Section~\ref{sec:training} zero-shot,
without fine-tuning. The operator issues only a constant forward
command ($\approx$1.0~m/s) with no per-terrain switching or
adjustment. Indoor tests run under overhead fall protection.

\begin{figure}[!t]
\centering
\includegraphics[width=\columnwidth]{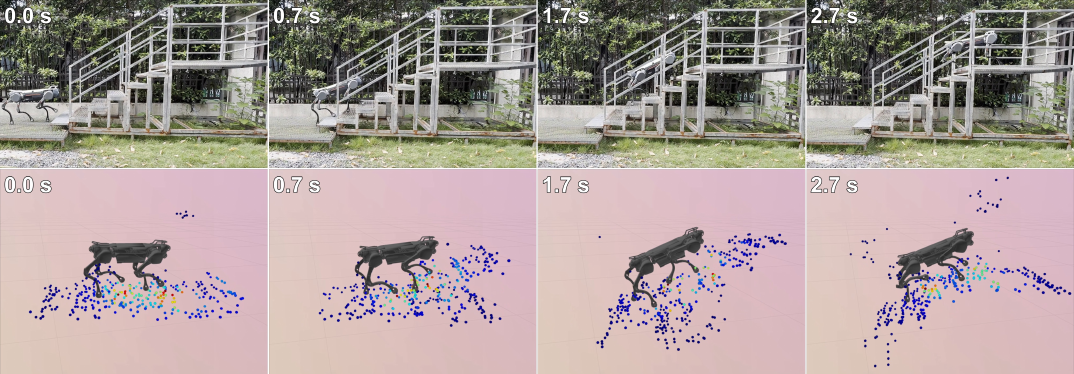}
\caption{Ascent of the steel-grating stairs (top) and the
simultaneous point-cloud attention (bottom; rows time-aligned, red =
high, blue = low): attention concentrates on the support under the
stance feet and the tread about to be stepped on; the sparse points
preserve the step structure with no map.}
\label{fig:stairpc}
\end{figure}

\begin{figure}[!t]
\centering
\includegraphics[width=\columnwidth]{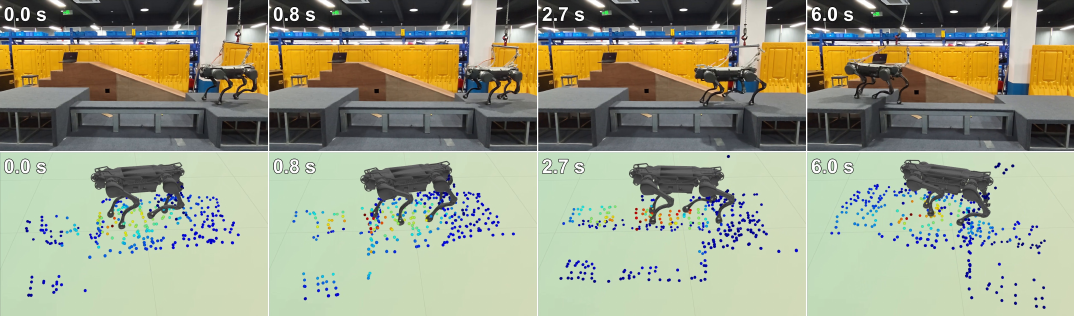}
\caption{Balance-beam crossing with the simultaneous point-cloud
attention (rows time-aligned; red = high, blue = low): attention
concentrates on the beam under the feet and extends to the next
segment to be stepped on; the sparse points preserve the narrow strip
with no map.}
\label{fig:beampc}
\end{figure}

\textbf{Terrain traversal.} The same policy passes six scenarios:
industrial steel-grating stairs (25-cm
steps, 45$^{\circ}$), natural outdoor slopes
(30$^{\circ}$--40$^{\circ}$), a 0.8-m gap, a 0.8-m platform, stepping
stones (35-cm tops, 30-cm spacing), and a balance beam (35-cm width).
The steepest slopes exceed the hardest curriculum level
(Table~\ref{tab:terrain}), an out-of-distribution crossing. The gap
and platform sit in the upper difficulty range of the two dynamic
incentives (Section~\ref{sec:reward}), and the stones match the
narrowest level, the beam approaching it. The grating stairs are the
most direct evidence for the raw-point interface
(Fig.~\ref{fig:stairpc}): the real treads are the see-through
structure that 2.5-D height maps cannot represent correctly, yet the
policy lands on solid
support straight from the points. The perception-to-foothold link
is set by the dual-channel simulation (Section~\ref{sec:training}). The
balance beam
(Fig.~\ref{fig:beampc}) is the sparse-foothold counterpart, crossed
with no map and no foothold estimate. In both figures, the attention
read out with the action (Section~\ref{sec:training}) concentrates
on the support beneath the feet and the next landing surface,
corroborating the proprioception-conditioned fusion
(Section~\ref{sec:policy}).

\begin{figure}[!t]
\centering
\includegraphics[width=\columnwidth]{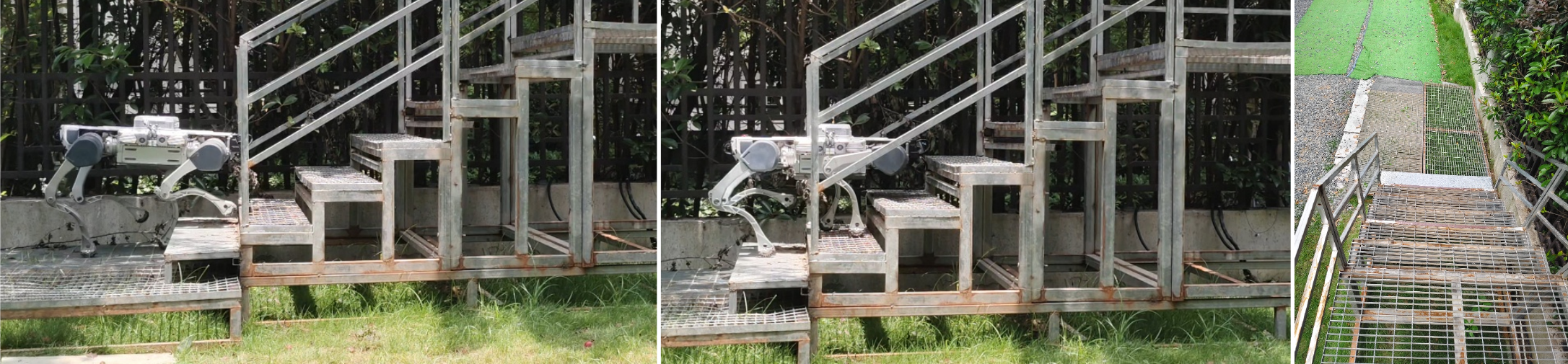}
\caption{Failure of the w/o Dual-channel terrain variant
(Section~\ref{sec:setup}) on the same stairs. Left: a front leg reaches
the first tread but the robot cannot mount; center: stalled at the stair
base, torso pitched up; right: top view of the see-through treads
(PointWalker: Fig.~\ref{fig:stairpc}).}
\label{fig:stairfail}
\end{figure}

\begin{figure}[!t]
\centering
\includegraphics[width=\columnwidth]{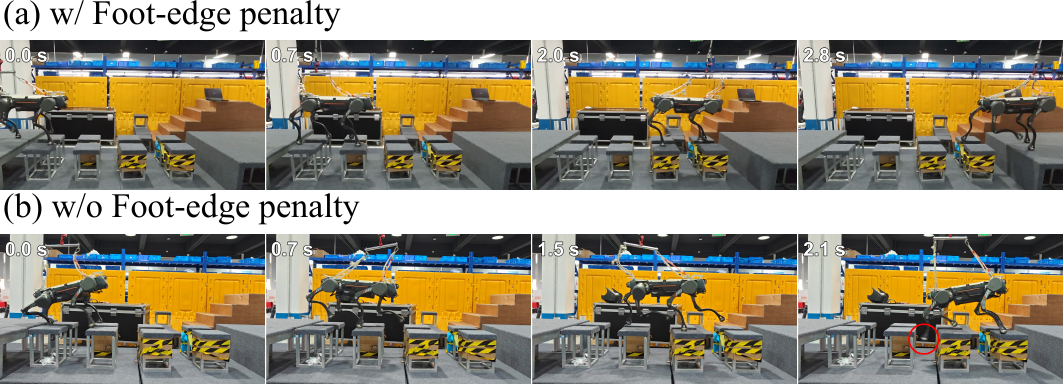}
\caption{Foot-edge penalty, same-site contrast on stepping stones.
\textbf{(a)}~PointWalker (w/ Foot-edge penalty): feet land inside the
35-cm stone tops. \textbf{(b)}~w/o Foot-edge penalty: a front foot
slips off a stone edge (red circle) and the robot falls.}
\label{fig:stepabl}
\end{figure}

\begin{figure}[!t]
\centering
\includegraphics[width=\columnwidth]{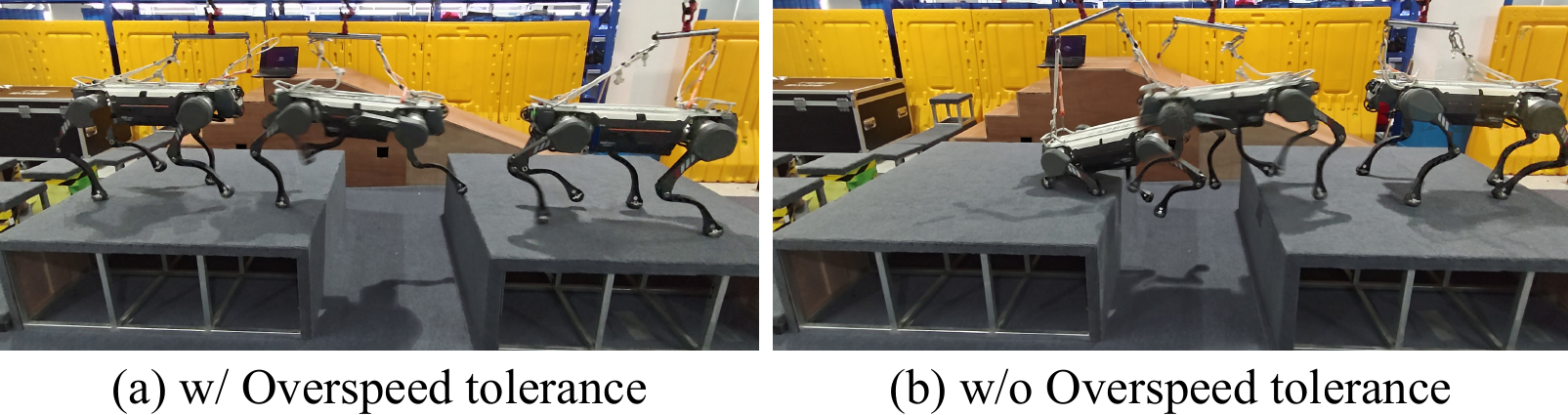}
\caption{Overspeed tolerance, same-site contrast at the same 0.8-m gap
(multi-exposure composites).
\textbf{(a)}~PointWalker (w/ Overspeed tolerance): accelerates before
the near edge and lands on the far side. \textbf{(b)}~w/o Overspeed
tolerance: momentum falls short; the nose strikes the far edge and
the robot hangs.}
\label{fig:gapcmp}
\end{figure}

\begin{figure}[!t]
\centering
\includegraphics[width=\columnwidth]{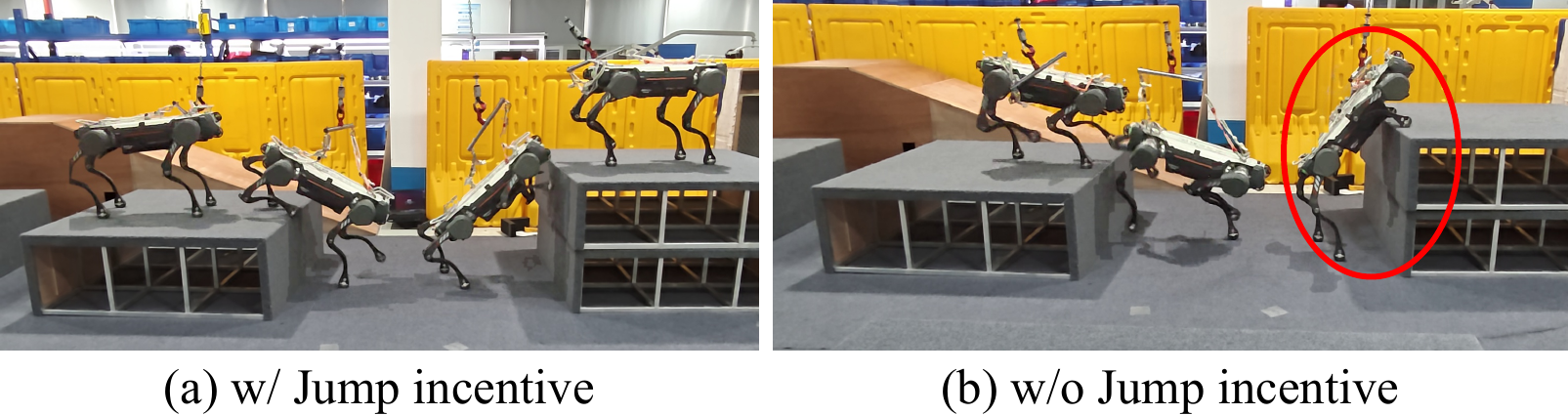}
\caption{Jump incentive, same-site contrast (multi-exposure
composites; descending from a single-tier platform and climbing the
0.8-m platform). \textbf{(a)}~PointWalker (w/ Jump incentive): hooks
the top edge and pulls itself onto the top. \textbf{(b)}~w/o Jump
incentive: claws at the top edge and stalls (red circle).}
\label{fig:jumpcmp}
\end{figure}

\textbf{Ablation verification.} Each variant fails at exactly the
maneuver its component supports. w/o Dual-channel terrain
saw only solid stairs in training. The hollow grating lies outside
its perception distribution, and it stalls at the stair base, at
most resting a front foot on the first tread (Fig.~\ref{fig:stairfail}),
where PointWalker completes the climb
(Fig.~\ref{fig:stairpc}). w/o
Foot-edge penalty slips off a stone edge and falls
(Fig.~\ref{fig:stepabl}), matching its threefold edge-landing ratio
(Fig.~\ref{fig:edge}). w/o Overspeed tolerance cannot build takeoff
momentum and nose-dives onto the far edge (Fig.~\ref{fig:gapcmp});
w/o Jump incentive claws at the top edge and never mounts
(Fig.~\ref{fig:jumpcmp}).

\begin{figure}[!t]
\centering
\includegraphics[width=0.88\columnwidth]{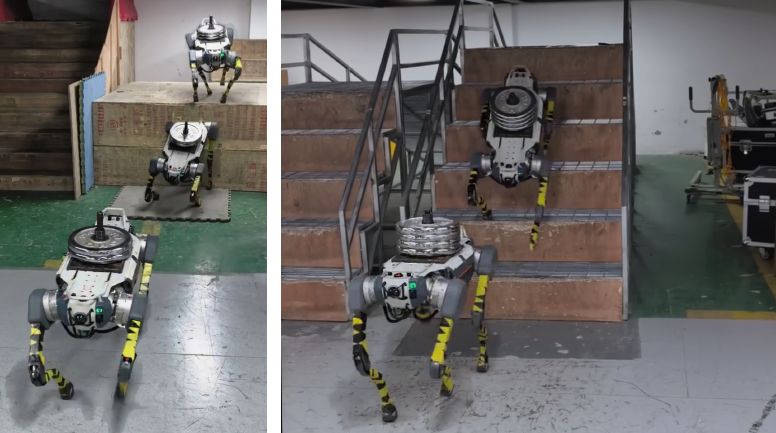}
\caption{Payload tests with rigid backpack loads (multi-exposure
composites): left, 30~kg (55\% of body weight) onto the 0.8-m
platform; right, 50~kg (91\%) up the grating stairs.}
\label{fig:payload}
\end{figure}

\textbf{Payload.} With rigid backpack loads, the X30 mounts the
0.8-m platform carrying 30~kg (55\% of body weight) and climbs the
grating stairs carrying 50~kg (91\%) (Fig.~\ref{fig:payload}). The
50-kg load equals the upper bound of the randomized payload range;
negative values emulate unloading (Table~\ref{tab:hyper}).

\begin{figure}[!t]
\centering
\includegraphics[width=\columnwidth]{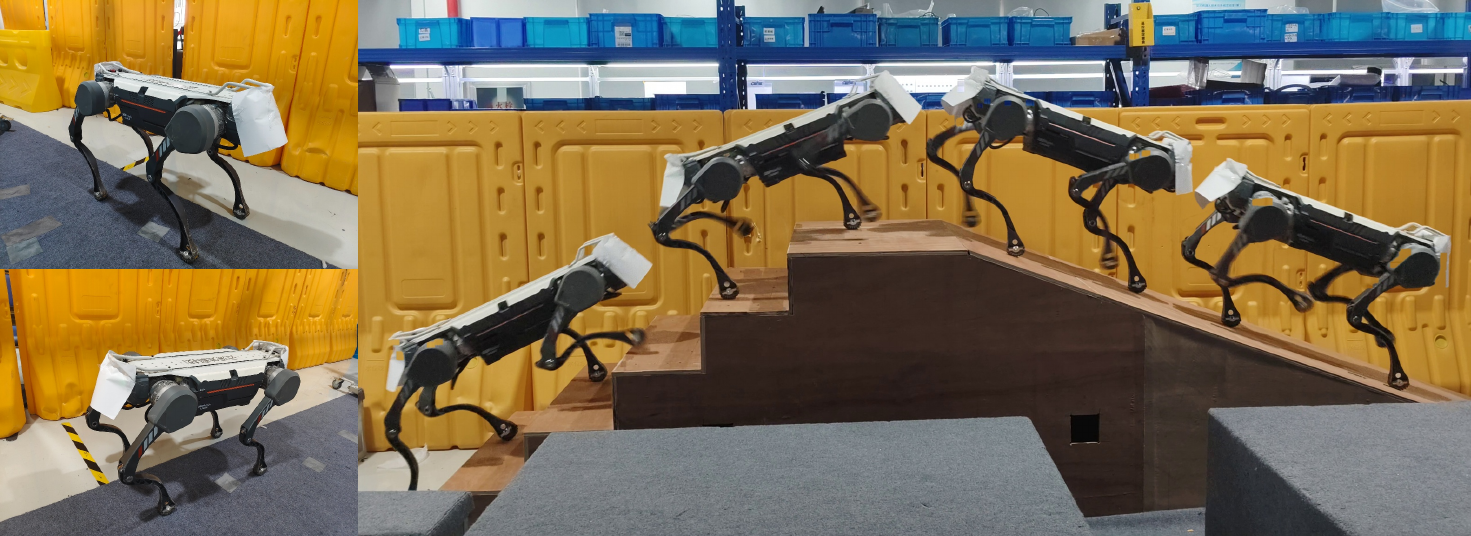}
\caption{LiDAR occlusion test. Top-left and bottom-left: one front
and one rear LiDAR blocked by adhesive covers (two in total). Right:
multi-exposure traverse of stairs (20-cm steps), platform, and
20$^{\circ}$ slope under occlusion.}
\label{fig:occl}
\end{figure}

\textbf{Occlusion robustness.} We block one front and one rear LiDAR
(Fig.~\ref{fig:occl}), removing far more returns than the 2\%
invalid-point degradation of training (Section~\ref{sec:perception}).
The robot still passes the combined scene of stairs, platform, and
slope (Fig.~\ref{fig:occl}, right). The stairs and slope lie inside the
curriculum, so the challenge is sensing degradation, met by the
LiDARs' field-of-view redundancy and the injected degradation.

\begin{figure}[!t]
\centering
\includegraphics[width=\columnwidth]{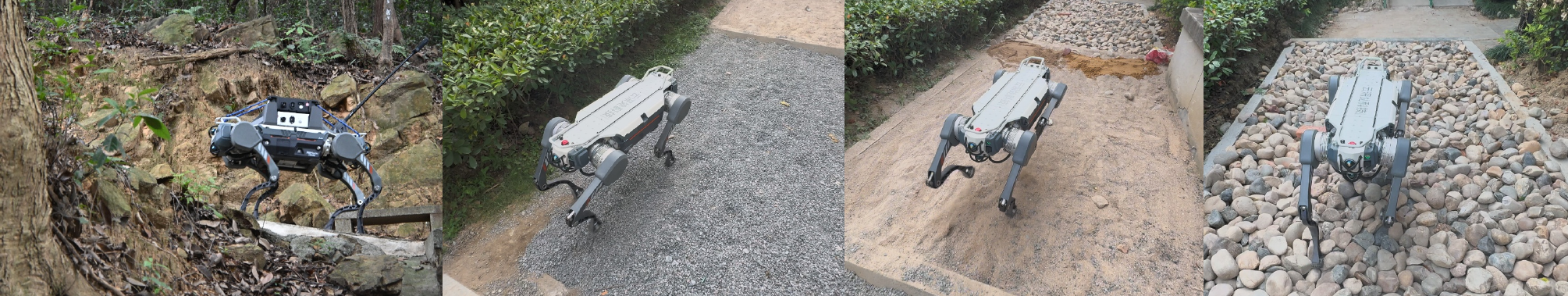}
\caption{Long-range outdoor and natural-terrain tests. Left: rocky
section of the 2.5-km route (200-m gain, 25-kg payload); right:
gravel, sand, and cobblestone, none of the seven training terrains.}
\label{fig:outdoor}
\end{figure}

\textbf{Long-range and natural terrain.} We also drive the policy,
carrying 25~kg (45\%), through a 2.5-km mountain route of 200~m
elevation gain in about 46~min at 0.9~m/s. The route crosses rock
slopes, leaf litter, and stone steps, and the policy runs
unadjusted throughout. Gravel, sand,
and cobblestone (Fig.~\ref{fig:outdoor}) lie outside the seven
training classes and differ in contact behavior (loose, sinking,
rounded). The same policy walks stably on all three without targeted
training, so its foothold decisions transfer beyond the training
domain. Reported
hardware gap crossings reach 0.8~m and platform jumps
0.5~m~\cite{cheng2024extreme}
under differing platforms and protocols (not directly comparable);
PointWalker crosses 0.8-m gaps and mounts 0.8-m platforms with one
mapless policy. All hardware experiments are in the supplementary
video.

\section{Discussion and Conclusion}
\label{sec:conclusion}

In this paper, we presented PointWalker, a mapless framework in
which one PPO run trains a single policy that attends directly to
raw LiDAR points across seven terrain classes. In simulation,
intermediate representations degraded on sparse and see-through
support: height maps aggregated the grating into spurious heights,
and depth images never saw beneath or beside the body. The raw
points preserved that geometry. Only a small fraction of them are
decisive footholds, so replacing cross-attention with mean pooling
cost 10 points of success rate; linear attention held the perception
pathway at 2.01~ms, a tenth of the control period. On hardware,
each ablated training-only design was essential: without
dual-channel simulation, the policy stalled at the stair base, and
without the foot-edge penalty, it slipped off the stones. Without
the traversal incentives, it reached the gap short of momentum and
the platform without a leap. Zero-shot on the X30, the same policy
crossed 0.8-m gaps, mounted 0.8-m platforms, walked on with two of
four LiDARs blocked, and completed a 2.5-km mountain route, all
without a terrain map.

Perception is local and momentary, so
the policy traverses only
what the current points show; porting beyond the X30's four-LiDAR
layout is untested. Within that scope, it generalized to
natural terrain beyond the training classes; global path planning
and moving support remain open. The first extension embeds
PointWalker in a navigation stack with a global planner. Another
is dynamic environments, extending the interface to moving
obstacles and support.

\bibliographystyle{IEEEtran}
\def\IEEEbibitemsep{0pt plus .5pt minus 1pt}
\bibliography{refs}

\end{document}